\documentclass[letterpaper]{article}
\usepackage{aaai}
\usepackage{times}
\usepackage{helvet}
\usepackage{courier}
\usepackage{xspace}
\usepackage{amssymb}
\usepackage{epic}
\usepackage{graphicx}
\usepackage{subfigure}
\usepackage{epic}
\usepackage{latexsym}
\usepackage{rotating}
\usepackage{multirow}

\newcommand{\nina}[1]{{#1}}

\begin{document}
\title{Eliminating the Weakest Link: Making Manipulation Intractable?}
\author{
Jessica Davies\\ University of Toronto\\ Toronto, Canada\\ jdavies@cs.toronto.edu  \And
Nina Narodytska\\NICTA and UNSW\\ Sydney, Australia\\ ninan@cse.unsw.edu.au \And
Toby Walsh\\ NICTA and UNSW\\ Sydney, Australia\\ toby.walsh@nicta.com.au}

\maketitle
\begin{abstract}
Successive elimination of candidates is often a route to making manipulation intractable to compute.
%We show that successively
%eliminating candidates is often a route to
%making manipulation intractable
%to compute. 
We prove that eliminating candidates does not
necessarily increase the computational complexity
of manipulation. However, for many voting rules used in
practice, the computational complexity increases.
For example, it is already known that it
is NP-hard to compute how a single voter
can manipulate the result of single transferable voting(the
elimination version of plurality voting).
We show here that it
is NP-hard to compute how a single voter
can manipulate the result of the elimination version
of veto voting, of the closely related Coombs' rule,
and of the elimination versions of a general class of
scoring rules.
\end{abstract}

\newtheorem{mydefinition}{Definition}
\newtheorem{mytheorem}{Theorem}
\newtheorem{myproposition}{Proposition}
\newtheorem{mylemma}{Lemma}
\newtheorem{myexample}{Example}
%\newtheorem{mytheorem}{Observation}
%\spnewtheorem*{myexample}{Running Example}{\bf}{\it}
%\newtheorem*{myexample}{Running Example}{\bf}{\it}
%\newtheorem{myexample}{}{\bf}{\it}
%\newtheorem*{myexampletwo}{Example}{\bf}{\it}
%\newtheorem*{myexample}{Running Example}
\newtheorem{mytheorem1}{Theorem}
\newcommand{\myproof}{\noindent {\bf Proof:\ \ }}
\newcommand{\myqed}{\mbox{$\blacksquare$}}
\newcommand{\myend}{\mbox{$\clubsuit$}}

\newcommand{\mymod}{\mbox{\rm mod}}
\newcommand{\mymin}{\mbox{\rm min}}
\newcommand{\mymax}{\mbox{\rm max}}
\newcommand{\range}{\mbox{\sc Range}}
\newcommand{\roots}{\mbox{\sc Roots}}
\newcommand{\myiff}{\mbox{\rm iff}}
\newcommand{\alldifferent}{\mbox{\sc AllDifferent}}
\newcommand{\permutation}{\mbox{\sc Permutation}}
\newcommand{\pecedence}{\mbox{\sc Perecendece}}
\newcommand{\disjoint}{\mbox{\sc Disjoint}}
\newcommand{\cardpath}{\mbox{\sc CardPath}}
\newcommand{\CARDPATH}{\mbox{\sc CardPath}}
\newcommand{\common}{\mbox{\sc Common}}
\newcommand{\uses}{\mbox{\sc Uses}}
\newcommand{\lex}{\mbox{\sc Lex}}
\newcommand{\usedby}{\mbox{\sc UsedBy}}
\newcommand{\nvalue}{\mbox{\sc NValue}}
\newcommand{\slide}{\mbox{\sc CardPath}}
\newcommand{\sliden}{\mbox{\sc AllPath}}
\newcommand{\SLIDE}{\mbox{\sc CardPath}}
\newcommand{\circularslide}{\mbox{\sc CardPath}_{\rm O}}
\newcommand{\among}{\mbox{\sc Among}}
\newcommand{\mysum}{\mbox{\sc MySum}}
\newcommand{\amongseq}{\mbox{\sc AmongSeq}}
\newcommand{\atmost}{\mbox{\sc AtMost}}
\newcommand{\atleast}{\mbox{\sc AtLeast}}
\newcommand{\element}{\mbox{\sc Element}}
\newcommand{\gcc}{\mbox{\sc Gcc}}
\newcommand{\gsc}{\mbox{\sc Gsc}}
\newcommand{\contiguity}{\mbox{\sc Contiguity}}
\newcommand{\PRECEDENCE}{\mbox{\sc Precedence}}
\newcommand{\assignnvalues}{\mbox{\sc Assign\&NValues}}
\newcommand{\linksettobooleans}{\mbox{\sc LinkSet2Booleans}}
\newcommand{\domain}{\mbox{\sc Domain}}
\newcommand{\symalldiff}{\mbox{\sc SymAllDiff}}
\newcommand{\alldiff}{\mbox{\sc AllDiff}}
\newcommand{\doublelex}{\mbox{\sc DoubleLex}}
\newcommand{\myldots}{\mbox{..}}
\newcommand{\myscore}{\mbox{$\sigma$}}
\newcommand{\truncated}{\mbox{$score_t$}}

%%%%%%%%%%%%%
\newcommand{\dyn}{\mbox{\sc Dyn}}
\newcommand{\preced}{\mbox{\sc Prec}}
\newcommand{\prep}{\mbox{\sc Prep}}
\newcommand{\precedsh}{\mbox{\sc $\preced_{sh}$}}
\newcommand{\precedprep}{\mbox{\sc $\preced+\prep$}}
\newcommand{\precedprepsh}{\mbox{\sc $\preced+\prep_{sh}$}}

\newcommand{\slidingsum}{\mbox{\sc SlidingSum}}
\newcommand{\MaxIndex}{\mbox{\sc MaxIndex}}
\newcommand{\REGULAR}{\mbox{\sc Regular}}
\newcommand{\regular}{\mbox{\sc Regular}}
\newcommand{\precedence}{\mbox{\sc Precedence}}
\newcommand{\STRETCH}{\mbox{\sc Stretch}}
\newcommand{\SLIDEOR}{\mbox{\sc SlideOr}}
\newcommand{\NAE}{\mbox{\sc NotAllEqual}}
\newcommand{\mytheta}{\mbox{$\theta_1$}}
\newcommand{\mysigma}{\mbox{$\sigma_2$}}
\newcommand{\mysigmatwo}{\mbox{$\sigma_1$}}

\newcommand{\todo}[1]{{\tt (... #1 ...)}}
\newcommand{\myOmit}[1]{}

\newcommand{\dpsb}{DPSB}
\newcommand{\br}[1]{\mbox{$\bar{ #1}$}}

\section{Introduction}

Voting is a general mechanism for combining the
preferences together of multiple agents.
Voting is, however, not without its problems.
One such problem is that agents may vote strategically,
mis-reporting their true preferences in order to
improve the outcome for them.
For instance, in each round of a popular TV game show,
players vote on which other player to eliminate.
The host, Anne Robinson then tells the unlucky player,
``You are the {\em weakest} link, goodbye!''.
Players have an interesting strategic
decision to make. On the one hand, they
should vote to eliminate weak players
(as weak players will tend to reduce the
size of the jackpot). On the other hand, they
should vote to eliminate strong players
(as the overall winner takes the final
jackpot and everyone else walks away empty-handed).
Similarly, when the International Olympic Committee (IOC)
meets to select a site for the next Olympics, there is
an election in which the weakest city is
successively eliminated. Strategic voting
often appears to take place. % in this election.
For example, in the vote
for the site of the 2012 Olympics, New York
had 19 votes in the first round but
only 16 in the second as several IOC members
switched allegiances.
In this paper, we study the computational resistance
of elimination style voting rules to such strategic
voting.

Results like those of Gibbard-Satterthwaite
prove that most voting rules are manipulable. That is, it may pay
for agents to mis-report their preferences.
One potentially appealing escape from manipulation
is computational complexity
\cite{bartholditoveytrick}.
Whilst manipulations might exist, what if
they are too hard to find?
Unfortunately, only a few voting rules
used in practice
are known to be NP-hard to manipulate
with unweighted votes and a single manipulator:
single transferable voting (STV)~\cite{stvhard},
a variant of the Copeland rule~\cite{bartholditoveytrick},
ranked pairs~\cite{xzpcrijcai09},
and Nanson's and Baldwin's rules
\cite{nwxaaai11}.
A feature common to a majority of
these rules is that they successively
eliminate candidates.
We therefore explore
in more detail whether such elimination style
voting makes manipulation intractable to compute.

\section{Background}

We consider a general class of voting
rules. A {\em scoring rule} over $m$ candidates is defined by a
vector $(s_1,\ldots,s_m)$ where for
each vote ranking a candidate in position
$i$, the candidate receives a score of $s_i$.
The candidate with the highest total score wins.
Plurality has the scoring vector
$(1,0,\ldots,0)$,
Borda has $(m-1,m-2,\ldots,0)$,
whilst veto has $(1,\ldots,1,0)$.
For a scoring rule $X$ with scoring vector
$(s_1,\ldots,s_m)$, the {\em adjoint} of $X$, written $X^*$ 
has the scoring vector
$(s_1-s_m,\ldots,s_1-s_2,s_1-s_1)$. % \cite{ctcats2007}.
For example, the adjoint of plurality is
veto. %, whilst the adjoint of Borda is Borda itself. 
Note that $(X^*)^* = X$.

Elimination versions of scoring rules can vary
along a number of dimensions:
\begin{description}
\item[Base rule:] STV is
an elimination version of plurality voting,
whilst Nanson's and Baldwin's rules
are elimination versions of Borda voting.
We consider here elimination versions
of other scoring rules like
veto voting.
\item[Elimination criteria:]
Different criteria can be used to eliminate
candidates. For instance, in STV and Baldwin's
rule, we succcessively eliminate the last
placed candidate. On the other hand, in
Nanson's rule, we eliminate all candidates
with less than the average Borda score.
\item[Stopping criteria:]
Do we stop when all but one candidate has
been eliminated, or as soon as
one candidate has a majority of first placed
votes? For example, Coombs' rule is an elimination
version of veto voting which stops when one candidates 
has a majority.
\item[Voting:]
Do agents vote just once, or 
in each round? For example, in STV
voting, agents vote only once. On the other
hand, when selecting Olympic venues, IOC
members can cast a new vote in each round.
We shall show
that this increases the opportunity for manipulation.
\end{description}

Given a voting rule $X$, 
$eliminate(X)$ is the rule
that successively eliminates the candidate
placed in last place by $X$.
For a scoring rule $X$, $divide(X)$ is the rule
that successively eliminates those
candidates with the mean or smaller score.
For non-scoring rules $X$, $divide(X)$ is the rule
that successively eliminates candidates ranked by $X$ 
in the bottom half. Finally, $sequential(X)$
is the voting rule which runs a
sequence of elections using $X$ to
eliminate the last placed candidate from
each successive election. In each round, % of $sequential(X)$, 
voters can change their
vote according to which candidates remain.

\begin{myexample}
STV is $eliminate(plurality)$.
Note that $eliminate(STV)$ is STV itself.
Baldwin's rule is $eliminate(Borda)$.
Nanson's rule is $divide(Borda)$.
Exhaustive ballot is $sequential(plurality)$.
The IOC uses $sequential(plurality)$ to
select Olympic sites.
The FIFA executive committee
uses the same rule to select
the location of the World Cup.
The TV game shows, ``Survivor'' and ``The Weakest Link''
both use $sequential(veto)$ to eliminate
players up to the final round.
%Note that 
Finally, Coombs' rule is related
to $eliminate(veto)$. %and is, in fact, sometimes confused for it. 
Coombs' rule successively eliminates
the candidate in last place in the
most votes until there is a candidate with
a majority of first place votes.
\end{myexample}

Elimination style voting rules satisfy several
desirable axiomatic properties. For example,
consider Condorcet consistency, the property that
a voting rule elects the candidate that beats 
all others in pairwise comparisons when such a candidate
exists. Whilst the Borda rule is not Condorcet consistent,
elimination versions of Borda voting like Nanson's and Baldwin's 
rule are Condorcet consistent. On the other hand,
elimination rounds can also destroy a deriable
axiomatic property. In particular, consider monotonicity, 
the property that raising the position of the winner
in some ballots does not change the winner. 
Whilst Borda voting is monotonic,
elimination style voting rules like STV, Nanson's 
and Baldwin's are not monotonic. 
The loss of monotonicity is one of the 
significant trade-offs involved in
obtaining a voting rule that is, as we shall
see, somewhat more resistant to manipulation.

\section{Manipulation}

Successively eliminating candidates can
increase the complexity of computing a
manipulation. 
For example, computing a manipulation of
plurality %by one manipulator
is polynomial, but of $eliminate(plurality)$
is NP-hard \cite{stvhard}. Similarly, computing a manipulation of
Borda by one manipulator
is polynomial, but of $eliminate(Borda)$ and
$divide(Borda)$ is NP-hard \cite{nwxaaai11}.
Elkind and Lipmaa (\citeyear{elisaac05})
conjectured that many elimination
style voting rules will be %computationally
intractable to manipulate. 
They argue that ``{\em [such elimination style]
protocols provide the most
faithful manipulation-resistant approximation to the
underlying protocols, which makes them compelling alternatives
to the original protocols}''.

We might wonder if elimination always
increases tie computational complexity.
The following result demonstrates that
it does not always make computing a manipulation intractable.

\begin{mytheorem}
There exists a non-dictatorial voting rule $X$
for which computing a manipulation of $X$,
$eliminate(X)$ and $divide(X)$ are polynomial.
\end{mytheorem}
\myproof
Consider the %unanimity or alphabetical 
rule which orders candidates alphabetically
unless there is unanimity when it returns the
unanimous order.
% Computing a manipulation
%of $X$ is polynomial. If the manipulators
%want the alphabetically least candidate to
%win, they put this candidate in first
%place and the other candidates in any other
%order. Otherwise, if there is unanimity amongst
%the non-manipulators and the manipulators want
%the most preferred candidate in this vote
%to win, then they vote as the non-manipulators.
%In all other cases, the manipulators
%cannot make their most preferred candidate win.
%Computing a manipulation of
%$eliminate(X)$ is also polynomial.
%There are two cases.
%In the first case, we can
%eliminate some number of candidates
%from the end of the alphabet to give
%an unanimous vote from the non-manipulators. If
%the manipulators want
%the most preferred candidate in this vote
%to win, then they vote as one of the non-manipulators.
%Otherwise, the only candidate that can
%win is the alphabetically least candidate
%and it does not matter how the manipulators vote.
%Computing a manipulation of $divide(X)$ is
%polynomial for similar reasons.
\myqed

Indeed there are even (admittedly artificial)
voting rules where successively
eliminating candidates reduces the computational
complexity of computing a manipulation.

\begin{mytheorem}
There exists a voting rule $X$
for which computing a manipulation of $X$ is NP-hard
but of $eliminate(X)$ and of $divide(X)$ are polynomial.
\end{mytheorem}
\myproof
Let candidates be integers in $[0,m]$.
$X$ is a rule whose result decides
%the satisfiability of 
a 1-in-3-SAT problem on positive clauses
over $m$ %Boolean 
variables.
A vote which starts with %candidate 
0 is
interpreted as a positive clause
by taking the candidates in 2nd to 4th place
as its literals. 
Any other vote 
is interpreted as a truth 
assignment: those
candidates appearing before
0 are interpreted as true literals, and 
those after as false. With 2 candidates,
$X$ returns the majority winner. With 3 or
more candidates, $X$ returns
0 as winner if one of the votes
represents a truth assignment which satisfies exactly 1 in
3 literals in each clause represented by a vote,
otherwise 1 is winner.
Other candidates are returned in numerical order.
Computing a manipulation of $X$ is NP-hard.
However, computing a manipulation
of $eliminate(X)$ or $divide(X)$ is polynomial. 0
and 1 always enter the final round, and the
overall winner is simply the majority
winner between 0 and 1.
\myqed

\section{Eliminate(veto)}

Adding elimination
rounds to plurality makes finding a manipulation
intractable. 
Veto is essentially the opposite rule to plurality.
This is reflected in the alternate name
for veto of ``anti-plurality''.
Computing a manipulation of veto is polynomial. We just
veto the current winner until our chosen candidate
wins. An interesting case to consider
then is $eliminate(veto)$. 
With weighted votes, Coleman and Teague
(\citeyear{ctcats2007})
have proved that computing a manipulation of
$eliminate(veto)$ is polynomial when the number of
candidates is bounded\footnote{Note that
Coleman and Teague call the voting rule studied
in their paper Coombs' rule but it is, in fact, $eliminate(veto)$.}.
They left 
``{\em 
the difficult[y]
of WCM [weighted coalition manipulation] 
on Coombs for unlimited candidates as an
open question}''.
We resolve this open question.
Computing a manipulation of $eliminate(veto)$
and of the closely related Coombs' rule
is NP-hard even with {\em unweighted} votes
and an unbounded number of candidates.

\newcommand{\sr}{\mbox{\sc ShortScoring}}
\newcommand{\elimsr}{\mbox{\sc eliminate(ShortScoring)}}
\newcommand{\divsr}{\mbox{\sc divide(ShortScoring)}}
\newcommand{\seqsr}{\mbox{\sc sequential(ShortScoring)}}
\newcommand{\elimapp}{\mbox{\sc Eliminate(k-approval) }}
\newcommand{\elimveto}{\mbox{\sc Eliminate(veto)  }}

\begin{mytheorem}
		\label{l:l1}
        Deciding if a single manipulator can make a candidate
win for $eliminate(veto)$ is NP-complete.
		\end{mytheorem}		
		\myproof (Sketch, the full proof can be found online in a technical report).
        The proof is inspired by ideas from~\cite{stvhard}. We reduce from the 3-COVER problem. We are given a set $S = \{d_1,\ldots,d_n\}$ with $|S| = n$ and
        subsets $S_1, S_2, \ldots,  S_m \subset S$ with $|S_i| = 3$ for $i\in[1,m]$.
        We ask if there exists an index set $I$ with $|I| = n/3$ and
        $\bigcup_{i\in I} S_i = S$. This set of $S_i$ is called a cover for $S$.
We create an \emph{eliminate(veto)} election
        such that a manipulator can make a given candidate win
iff there exists
        a cover for $S$.
        
        The set of all candidates is $C$ where $|C|=c$, and consists of 7 groups:
%        \begin{itemize}
`preferred candidate' $p$;
`items' $\{d_1,\ldots, d_n\}$ and an extra `item' $d_0$;
`first losers' $\{a_1,\bar{a}_1\ldots, a_m,\bar{a}_m\}$;
`second line' $\{b_1,\bar{b}_1,\ldots, b_m, \bar{b}_m\}$ ; %Each $b_i$ is associated with $S_i$, $i=1,\ldots,m$.
`pumps' $\{p_1,\ldots,p_m\}$;
`switches' $s_1$ and $ s_2$;
`garbage collectors' $g_{d_0}, \ldots, g'_{s_1} $ (see Table~\ref{t:t1} for the complete  list of garbage collectors).
 %       \end{itemize}
The elimination has 4 phases: $(1)$ cover selection, $(2)$,$(3)$
 cover verification $(4)$ garbage collection.

\begin{table}[!t]
\hspace{-1.5em}
\scriptsize{
\begin{tabular}{|@{}c@{}|@{}c@{}@{}c@{}||@{}c@{}||@{}r@{}r@{}c@{}c@{}l@{}l@{}|}
          \hline
            $\#$& & & $\#$ votes &\multicolumn{6}{|c|}{ Type of votes } \\
            % \cline{5-10}
           %     & & & & $c \prec $ & $c-1\prec$ & $c-2$& \myldots & $2$& $\prec1$\\
          \hline
                  \multicolumn{10}{|c|}{ \textbf{Block} $P_1$ }\\
                   \hline
                   \multicolumn{10}{|c|}{`preferred candidate' and `items' }\\
                   \hline
          1.& && $X-1$ & $p \prec$ & $\mathbf{g_p} \prec$ & & \myldots && $\prec\mathbf{g'_p}$ \\

         2.& & \multirow{1}{*}{} & $X-f_4 $ & $d_0 \prec$ & $\mathbf{g_{d_0}} \prec$ & & \myldots &$p$&$\prec\mathbf{g'_{d_0}}$ \\
         3.& $i\in [1,n]$ &  \multirow{1}{*}{}& $X-3$   & $d_i \prec$  & $\mathbf{g_{d_i}} \prec$    &  &\myldots  &$p$&$\prec\ \mathbf{g'_{d_i}}$ \\
          \hline
 \multicolumn{10}{|c|}{`First losers' and `second line'}\\
 \hline
         4.&\multirow{1}{*}{$i \in [1,m]$}  & \multirow{2}{*}{$\left.\vphantom{\rule{1mm}{0.4cm}}\right\lbrace$ }& $X-6$   & $b_i \prec$  & $\mathbf{g_{b_i}} \prec$    &  & \myldots &$p$&  $\prec\ \mathbf{g'_{b_i}}$ \\
         5.& $j \in S_i$  &   & $2$          & $b_i \prec$  & $d_j   \prec$   & $\mathbf{g_{b_id_j}} \prec$ & \myldots &$p$& $\prec\ \mathbf{g'_{b_id_j}}$  \\
% & & && & & & && \\
         6.&\multirow{2}{*}{$i \in [1,m]$} & \multirow{2}{*}{$\left.\vphantom{\rule{1mm}{0.4cm}}\right\lbrace$ } & $X-2$   & $\bar{b}_i \prec$  & $\mathbf{g_{\bar{b}_i}} \prec$    & &  \myldots &$p$&  $\prec \mathbf{g'_{\bar{b}_i}}$ \\
         7.& & & $2$          & $\bar{b}_i \prec$  & $d_0   \prec$    & $\mathbf{g_{\bar{b}_i d_0}} \prec$ & \myldots  &$p$& $\prec\ \mathbf{g'_{\bar{b}_i d_0}}$ \\

         \multirow{2}{*}{8.}&    $i \in [1,m)$& & $X-f_{123}$   & $a_i   \prec$  & $\mathbf{g_{a_i}} \prec$    & & \myldots  &$p$ &$ \prec \mathbf{g'_{a_i}}$  \\
                   & & & $X-f_{12}$ & $a_m \prec$ & $\ \mathbf{g_{a_m}}\prec$ & & \myldots &$p$ &$ \prec \mathbf{g'_{a_m}}$ \\
          9.&\multirow{2}{*}{$i \in [1,m]$} &\multirow{2}{*}{$\left.\vphantom{\rule{1mm}{0.4cm}}\right\lbrace$ } & $f_1$   &  $a_i \prec$  & $b_i \prec$    & $\mathbf{g_{a_ib_i}} \prec$ &  \myldots &$p $& $ \prec \mathbf{g'_{a_ib_i}}$  \\
           10.&   &  & $f_2$   & $a_i \prec $  & $\bar{a}_i \prec $    & $p_i \prec$ &  $\mathbf{g_{a_i\bar{a}_i p_i}} \  \myldots\ $ &$p$& $\prec \mathbf{g'_{a_i\bar{a}_i p_i}}$ \\
           11.&  $i \in [1,m)$    & & $f_3$   & $a_i \prec $  & $\bar{a}_i \prec$    & $a_{i+1} \prec $ & $\mathbf{g_{a_i\bar{a_i}a_{i+1}}}  \  \myldots\ $ &$p$& $\ \prec \mathbf{g'_{a_i\bar{a}_ia_{i+1}}}$   \\
           12.& & & $f_3$ & $a_1 \prec $ & $\mathbf{g_{a_1}}\prec$ & & \myldots &$p$& $\prec \mathbf{g'_{a_1}}$ \\

          \multirow{2}{*}{13.} &      $i \in [1,m)$ && $X-f_{123}$   & $\bar{a}_i  \prec $  & $\mathbf{g_{\bar{a}_i}}\prec$    & & \myldots  &$p$& $ \prec  \mathbf{g'_{\bar{a}_i}}$  \\
            &         & & $X-f_{12}$   & $\bar{a}_m  \prec  $  & $\ \mathbf{g_{\bar{a}_m}}\prec$    & & \myldots  &$p$& $  \prec \mathbf{g'_{\bar{a}_m}}$  \\
          14. &  \multirow{2}{*}{$i \in [1,m]$} & \multirow{2}{*}{$\left.\vphantom{\rule{1mm}{0.4cm}}\right\lbrace$ } & $f_1$   & $\bar{a}_i  \prec $  & $\bar{b}_i\prec $    &  $\mathbf{g_{\bar{a}_i \bar{b}_i}}$ &  \myldots &$p$& $ \prec  \mathbf{g'_{\bar{a}_i \bar{b}_i}}$  \\
          15. & & & $f_2$   & $\bar{a}_i \prec $  & $a_i \prec $    & $p_{i} \prec $ &  $\mathbf{g_{\bar{a}_i a_ip_i}} \myldots$ &$p$&  $\prec  \mathbf{g'_{\bar{a}_i a_ip_i}}$  \\
          16. & $i \in [1,m)$& & $f_3$   & $\bar{a}_i \prec $  & $a_i \prec $    & $\bar{a}_{i+1} \prec $ &  $\mathbf{g_{\bar{a}_i a_i\bar{a}_{i+1}}} \myldots$ &$p$& $\prec \mathbf{g'_{\bar{a}_i a_i\bar{a}_{i+1}}} $   \\
          17. & & & $f_3$   & $\bar{a}_1 \prec  $  & $\mathbf{g_{\bar{a}_1}} \prec $      &  &  \myldots &$p$&$\prec  \mathbf{g'_{\bar{a}_1}} $    \\
 \hline
  \multicolumn{10}{|c|}{`Pumps'}\\
  \hline
          18. &   \multirow{4}{*}{ }   &\multirow{5}{*}{$\left.\vphantom{\rule{1mm}{1.1cm}}\right\lbrace$ }& $2f_2$   &  $p_i\prec $ & $a_j\prec $  & $ \mathbf{g_{p_i a_j}} $   & \myldots  &$p$& $\prec  \mathbf{g'_{p_i a_j}} $   \\
          19. & \multirow{2}{*}{ $i\in [1,m)$} &  & $2f_2$   &  $p_i\prec $ &  $\bar{a}_j\prec $  &  $ \mathbf{g_{p_i \bar{a}_j}} $   & \myldots  &$p$& $\prec  \mathbf{g'_{p_i \bar{a}_j}} $   \\
          20. &\multirow{2}{*}{$j \in (i,m]$}  &  & $2f_2$   &  $p_i\prec $ &${b}_j\prec $  &    $ \mathbf{g_{p_i {b}_j}} $   & \myldots  &$p$& $\prec  \mathbf{g'_{p_i {b}_j}} $   \\
          21. &      & & $2f_2$   &  $p_i\prec $ &$\bar{b}_j\prec $  &   $ \mathbf{g_{p_i \bar{b}_j}} $    & \myldots  &$p$& $\prec  \mathbf{g'_{p_i \bar{b}_j}} $   \\
          22. &      && $2f_2$   &  $p_i\prec $ &$p_j\prec $  &    $ \mathbf{g_{p_i {p}_j}} $   & \myldots  &$p$&  $\prec  \mathbf{g'_{p_i {p}_j}} $  \\
                \cline{3-10}
         23. &       &\multirow{5}{*}{$\left.\vphantom{\rule{1mm}{1.2cm}}\right\lbrace$ } & $2f_2$   &  $p_i\prec $ &  $p\prec $  &   $ \mathbf{g_{p_i {p}}} $  & \myldots  &&$ \prec  \mathbf{g'_{p_i {p}}} $    \\
         24. &     $i \in [1,m]$   & & $2f_2$   &  $p_i\prec $ &  $d_0\prec $  &   $ \mathbf{g_{p_i {d}_0}} $  & \myldots  &$p$&  $\prec  \mathbf{g'_{p_i {d}_0}} $   \\
         25. &    $k \in [1,n]$   & & $2f_2 $   & $p_i\prec $ &  $d_k \prec$  &    $ \mathbf{g_{p_i {d}_k}} $   & \myldots  &$p$& $\prec  \mathbf{g'_{p_i {d}_k}} $ \\
         26. &       & & $2f_2$   &  $p_i\prec $ &  $s_1 \prec $  &    $ \mathbf{g_{p_i {s}_1}}  \prec$   & \myldots  &$p$& $  \prec \mathbf{g'_{p_i {s}_1}} $   \\
         27. &       & & $f_1$   &  $p_m  \prec$ &  $s_1  \prec$  &  $ \mathbf{g_{p_m {s}_1}} $     & \myldots  &$p$&  $  \prec \mathbf{g'_{p_m {s}_1}} $  \\
         28. &       & & $2f_2$   &  $p_i  \prec$ &  $s_2  \prec$  &   $ \mathbf{g_{p_i {s}_2}} $    & \myldots  &$p$&   $  \prec \mathbf{g'_{p_i {s}_2}} $    \\
 \hline
   \multicolumn{10}{|c|}{`Switches'}\\
   \hline
        29. &  \multirow{3}{*}{ }   &\multirow{3}{*}{ }& $4f_2+f_1$     &  $s_1\prec$ &  $p\prec$  &   $ \mathbf{g_{s_1 p}} $   & \myldots  && $\prec \mathbf{g'_{s_1 p}} $   \\
        30. &        &   & $4f_2+f_1$   &  $s_1\prec$ &  $d_0\prec$  & $ \mathbf{g_{s_1 d_0}} $    & \myldots  &$p$&  $ \prec \mathbf{g'_{s_1 d_0}} $   \\
        31. &       $k \in [1,n]$  & & $4f_2+f_1$   &  $s_1\prec$ &  $d_k\prec$  & $ \mathbf{g_{s_1 d_k}} $    & \myldots  &$p$&$\prec \mathbf{g'_{s_1 d_k}} $     \\
        32. &        & & $4f_2+f_1$   &  $s_1\prec$ &  $s_2\prec$  & $ \mathbf{g_{s_1 d_k}} $    & \myldots  &$p$& $\prec \mathbf{g'_{s_1 d_k}} $    \\
% \hline
        33. &  \multirow{2}{*}{}   &\multirow{2}{*}{ }& $1$     &  $d_0\prec$ &  $s_2\prec$  &   $ \mathbf{g_{d_0 s_2}} $  & \myldots  & $p$ &  $\prec \mathbf{g'_{d_0 s_2}} $   \\
        34. &     $k \in [1,n]$    & & $1$   &  $d_k\prec$ &  $s_2\prec$  &   $ \mathbf{g_{d_k s_2}} $   & \myldots  &$p$& $\prec \mathbf{g'_{d_k s_2}} $   \\
        35. & & & $X-n-3$ & $s_2\prec$ & ${g_{d_0}}\prec$ & $g_{d_1}$ & \myldots && $\prec p$ \\
          \hline
           \multicolumn{10}{|c|}{ \textbf{Block} $P_2$ }\\
          \hline
        36. &    \multirow{1}{*}{$i \in [1,m]$}   &\multirow{1}{*}{ }& $X - \myscore_{P_1}^1(p_i)$   &  $p_i\prec$ &  $ \mathbf{g_{p_i}}\prec $   &     & \myldots  &$p$& $\prec \mathbf{g'_{p_i}}$   \\
        37. &     \multirow{1}{*}{}   &\multirow{1}{*}{}& $X - \myscore_{P_1}^1(s_1)$   &  $s_1\prec$ &  $ \mathbf{g_{s_1}}\prec $   &     & \myldots  &$p$& $\prec \mathbf{g'_{s_1}} $   \\
          \hline
% \multicolumn{10}{|c|}{ Block $P_3$ }\\
% \hline
%        38. &   \multirow{1}{*}{ $g_{y_iy_jy_k}$:}   &\multirow{1}{*}{ $\forall g_{y_iy_jy_k}$}& $X - \myscore_{P_1 \cup P_2}\left((y_iy_jy_k \myldots)\right)$   & ( $g_{y_iy_jy_k}$ &   &     & \myldots  &$c$& $g^+_{y_iy_jy_k}$)  \\
%        39. &   \multirow{1}{*}{ $g'_{y_iy_jy_k}$:}   &\multirow{1}{*}{ $\forall g'_{y_iy_jy_k}$}& $X$   & ( $g'_{y_iy_jy_k}$ &   &     & \myldots  &$c$& $g^+_{y_iy_jy_k}$)  \\
%        40. &   \multirow{1}{*}{ $g^+_{y_iy_jy_k}$:}   &\multirow{1}{*}{ $\forall g^+_{y_iy_jy_k}$}& $X$   & ( $g^+_{y_iy_jy_k}$ &   &     & \myldots  &$c$& $g'_{y_iy_jy_k}$)  \\
        \end{tabular}
        \caption{ The constructed election. }\label{t:t1}}
\end{table}
{Let  $\myscore^k(c')$, $c'\in C$ be the number of last place votes for candidate $c'$
at round $k$. We call this the veto-score of $c'$. }
{First, we explain the candidates. The candidates `first losers', `second line' and `pumps'
form a gadget to select a cover. 
There are $5m$ candidates of these types in total, logically partitioned
into elimination groups $\langle a_i, \bar{a}_i, b_i, \bar{b}_i, p_i \rangle$, $i\in[1,m]$.
The construction makes sure that $4$ out of $5$ elements of $i$th group are eliminated consecutively during 4 rounds starting at round $4i+1$,
$i \in [0,m)$.
Moreover, $\{a_i, \bar{a}_i, p_i\}$, $i\in[1,m]$ must be eliminated letting one of the $\{b_i, \bar{b}_i\}$
reach round $4m$. Eliminated candidates $b_i$ determine  selection sets $S_i$. 
The `pump' $p_i$  increases the veto-scores of all candidates except `garbage collectors'
and $i$ running candidates from groups $j$, $j \leq i$. This allows us to remember $i$ choices
of the manipulator encoded in these $i$ running candidates from  $2i$ candidates in $\{b_j, \bar{b}_j\}$, $j=1,\myldots,i$.
The `items' candidates encode the set of items.
The`switches' check the cover. 
%In particular,
 The first `switch' $s_1$ 
separates elimination of  $p,d_0,\ldots,d_n,s_2$ from the elimination of other candidates.
The second `switch' $s_2$
is the {\em most} dangerous candidate that can be eliminated iff a valid cover is selected during the first $4m$ rounds. `Garbage collectors' $g_{d_0}, \ldots, g_{s_1}$ control the
veto-scores of non-garbage candidates.
`Garbage collectors' $g'_{d_0}, \ldots, g'_{s_1}$ prevent $p$ from having a majority (which is needed later on to prove Theorem~\ref{l:l2}).
}

{ %Next we present votes.
We partition votes into two sets: $P_1$ and $P_2$.
Table~\ref{t:t1} shows the votes.
%We write all votes in the 
in \emph{reverse} preference order. % in the table.
We refer to sets of votes in each line of the table by the number in the third column.
For convenience,  we introduce a new garbage candidate in each set of votes. 
Unspecified candidates  are ordered in the same arbitrary order, starting with $g_{d_0},\ldots,g_{d_n}$ in all votes.
$P_1$ is the main construction. Lines 1--3 set up initial veto-scores for the preferred candidate and `items'.
Lines 4--22 encode the 1st phase.  
The important point to observe is how `pumps' work (lines 18--28). The candidate $p_i$ is eliminated
last in its group and increases the veto-score of all other running candidates by a constant $2f_2$ except
running candidates in $ \cup_{j=1}^i\{b_j, \bar{b}_j\}$ and `garbage' candidates. 
This allows  $m$ running candidates selected from $ \cup_{j=1}^m\{b_j, \bar{b}_j\}$ to reach the 4th phase. 
Lines 23--28 make sure that veto-score
of `items' and `switches', that are not eliminated during the 1st phase, grow the same way as veto-scores of $a$'s, $b$'s and $p$'s.
Line 27 is used to eliminate $s_1$ at round $4m+1$.
Lines 29-34 are used to check a cover. In particular, lines 33--34 are used to count
how many candidates $d_i$, $i\in[0,n]$ are eliminated by increasing veto-scores of  $s_2$.
Finally, line 35 is responsible
for  triggering the garbage collection procedure.
$P_2$ ensures that all `pumps'
and the `switch' $s_1$ have initial score of $X$,  where $X$ is sufficiently large number, e.g. $X > 16m^5$
and $\myscore^1_{P_1}(c')$ is the number of last place votes for candidate $c'$ in the votes $P_1$
at the first round.       
The initial veto-score of  a garbage collector equals $0$
and stays less than $X$ until the $4$th phase.
%is less than $X$ until the fourth phase.
%So while a non-garbage candidate has at least $X$ veto-points (such a candidate exists until the 4th phase), 
So we do not have to worry
about the garbage collectors during the first three phases. %in our arguments for the first three phases.
We also define the following constants required to control elimination
inside each group $\langle a_i, \bar{a}_i, b_i, \bar{b}_i, p_i \rangle$, $i \in [1,m]$.  The $f$ constants in  Table~\ref{t:t1} satisfy the following constraints:
        $f_{12} = f_1+f_2$, $f_{123}=f_1+f_2+f_3$,
        $f_1 \geq f_2 + 2f_3 + 2,$
        $2f_2 \geq f_1 + 2,$
        $2f_2 \geq f_3 + 2,$
        $f_1 \geq  f_3+2,$
        $f_i \geq 2m + 3$ for $i \in [1,3]$, and
        $f_4 = 2m-2n/3+3$.
        For example, $f_1 = 11(2m+3)$, $f_2=8(2m+3)$ and $f_3=3+(2m+3)$.
Overall, the construction ensures that initial veto-scores of all candidates $a$, $b$, $p$, $s_1$ equals $X$
with an exception of  $a_1$ and $\bar{a}_1$ that have $X+3$ veto-points.
All the other candidates have veto-scores that are less than or equal to $X$.
This forces the manipulator to make a choice between $a_1$ and $\bar{a}_1$ in the first round which triggers
a selection of sets in a cover.  
 We assume the tie-breaking rule: $s_2 \prec  d_0 \prec   p \prec  d_1,\prec  \ldots \prec  d_n, \ldots$.
%   Consider elimination phases.
}      
% We use these constants
%        to make the proof easier to follow.
%         Table~\ref{t:t2} shows the initial scores of all candidates (Round 0).     
%       \begin{itemize}
%         \item \textbf{Selection of sets. Rounds $1$ to $4m$.}  The key part of the first phase is the elimination of $m$ candidates, one from each pair $\{b_i,\bar{b}_i\}$, $i\in[1,m]$.  If $b_i$ is eliminated then the corresponding set $S_i$ is selected in the cover. A manipulator will choose which candidate from each pair is eliminated.
%         \item \textbf{Pump up of $\{p,d_0,\ldots,d_n\}$. Round $4m+1$.} This ensures that the veto-scores of $\{p,d_0,\ldots,d_n\}$ are greater than those of any other candidate by at least 2 points.
%        \item\textbf{Verifying the selected cover. Rounds $4m+2$ to $4m+2+(n+1)$.}  The third phase ensures  that $p$ wins iff
%        the number of eliminated candidates $b_i$ during the first phase is $n/3$ and covers $d_1,\ldots,d_n$.
%%        This  means that the corresponding $S_i$ form a cover of $S$.
%       \item \textbf{Collecting the garbage. Rounds $4m+2+(n+1)+1$ to $c$}. The final phase ensures that $p$ wins
%       if it is not already eliminated.
%       \end{itemize}

  %  Note that if a candidate $c'$ has a veto-score that is 2 or more than
  %  any other candidate then $c'$ is eliminated regardless of the manipulator's vote.
    %A proof that does not depend on tie-breaking is more complex.

  \textbf{Phase 1. Cover selection. Rounds $1$ to $4m$.}
     The 1st phase eliminates $m$ candidates, one from each pair $\{b_i,\bar{b}_i\}$, $i\in[1,m]$.  If $b_i$ is eliminated then 
     %the corresponding 
     the set $S_i$ is selected in the cover. A manipulator will choose which candidate from each pair is eliminated.
We claim the following holds in the first $4m$ rounds
where $i \in [0,m-1]$:
%\begin{itemize}
  %\item 
%\newline \indent  \textit{ $\triangleright$ Round $4i$.} The following invariant holds immediately before the $4i+1$st candidate is eliminated:
%      $$\myscore^{4i}(a_{i+1}) = \myscore^{4i}(\bar{a}_{i+1}) \geq \myscore^{4i}(y) + 3,$$
%      where $y \in C \setminus \{a_{1}, \bar{a}_{1}, \ldots, a_{i+1}, \bar{a}_{i+1},  b_1, \bar{b}_1, \ldots, b_{i}, \bar{b}_{i} \}$.
  %\item
\newline  \indent \textit{$\triangleright$ Round $4i +1$.}  
		The following invariant holds immediately before the $4i+1$st candidate is eliminated:
      $$\myscore^{4i+1}(a_{i+1}) = \myscore^{4i+1}(\bar{a}_{i+1}) \geq \myscore^{4i+1}(c') + 3,$$
      where $c' \in C \setminus \{a_{1}, \bar{a}_{1}, \myldots, a_{i+1}, \bar{a}_{i+1},  b_1, \bar{b}_1, \myldots, b_{i}, \bar{b}_{i} \}$.
  %\item
The manipulator can select which of $a_{i+1}$ or $\bar{a}_{i+1}$  is eliminated.
  %\item
  
   The manipulator \textbf{cannot} change the outcome of the following three rounds.
\newline \indent  \textit{$\triangleright$ Round $4i+2$.}
   $b_{i+1}$ is eliminated at this round iff
     ${a}_{i+1}$ is eliminated at the previous round. Similarly, $\bar{b}_{i+1}$ is eliminated iff
     $\bar{a}_{i+1}$ is eliminated at the previous round.
%      The manipulator \textbf{cannot} change the outcome of this round.
  %\item
\newline   \indent\textit{$\triangleright$ Round $4i+3$.}  $a_{i+1}$ is eliminated at this round iff
     $\bar{a}_{i+1}$ is eliminated at the $4i +1$st round. Similarly, $\bar{a}_{i+1}$ is eliminated at this round iff
     ${a}_{i+1}$ is eliminated at the $4i +1$st round. % The manipulator \textbf{cannot} change the outcome of this round.
  %\item
\newline   \indent \textit{$\triangleright$ Round $4i+4$.}  The candidate $p_{i+1}$  is eliminated. % at the $4i+4$th round.  
%The manipulator \textbf{cannot} change the outcome of this round.

%\end{itemize}

   Hence, the manipulator select $m$ candidates in $\cup_{i=1}^m\{b_i,\bar{b}_i\}$ to eliminate.
    The elimination of $p_m$ at round $4m$ forces an increase of the veto-scores of
    $p,d_0,\ldots,d_n,s_1,s_2$ by $2f_2$ (lines 23--26,28)  and an additional  increase of $f_1$ in the veto-score of $s_1$ (line 27).
    This means $s_1$ is the next candidate to be eliminated.
    
    \paragraph{ Phase 2. Pump up of $p,s_2, d_0,\ldots,d_n$. Round $4m+1$.}
     The elimination of $s_1$ increases the veto-scores of ${p,s_2,d_0,\ldots,d_n}$ by $4f_2+f_1$.
%      This ensures that the veto-scores of these candidates are larger than that of any other candidate. % by at least 2 veto-points.
      %The closest is $b_m$ or $\bar{b}_m$ with  veto-score $X + f_1 + 2f_2(m-1)$.
%     The resulting veto-scores after $4m+1$ rounds are shown at Table~\ref{t:t1} (Round $4m+1$).
    
    \paragraph{ Phase 3. Cover verification. Rounds $4m+2$ to $4m+2+(n+1)$.}
     This phase ensures  that $p$ reaches the next phase iff
     the sets $S_i$ that correspond to eliminated candidates $b_i$
     form a cover of $d_1,\ldots,d_n$ and there are exactly $n/3$ such candidates.
     %This  means that the corresponding $S_i$ form a cover of $S$.
      Consider the candidates ${p,s_2, d_0,\ldots,d_n}$. We observe that at round $4m+2$:
      $$\myscore^{4m+2}(p) = \myscore^{4m+2}(d_i) - 1 + 2 - y^{4m+2}_i,$$
      $$\myscore^{4m+2}(p) = \myscore^{4m+2}(d_0) - 1 + 2(m-n/3+1) - y^{4m+2}_0,$$
      $$\myscore^{4m+2}(p) = \myscore^{4m+2}(s_2) - 1 + (n+1) + 2,$$

        {where $y^{4m+2}_i$, $i \in [0,n]$ is the veto-score that candidate $d_i$ gets during the
        first $4m+1$ rounds in addition to its initial veto-score, $y^{4m+2}_i$ is even.}
      As can be seen from these equations, $s_2$ can be eliminated before $p$ iff $s_2$ gets
      $n+2$ extra veto-points. This is possible      
       iff
      $d_0,\ldots,d_n$ are eliminated before $s_2$ so that $s_2$ gets $n+1$ veto-points from lines 33--34.
      Moreover, the manipulator must give an extra veto-point to $s_2$.
      Then, by the tie-breaking rule, $s_2$ is eliminated before $p$.
      Consider how to eliminate  $d_0,\ldots,d_n$ before $s_2$ and $p$.
%      \begin{description}
        %\item[Candidates $d_i$ for $i\in [1,n]$:] 
       \newline  \indent  \textit{$\triangleright$ Candidates $d_i$ for $i=1,\ldots,n$:}
        Let $d_k$ be the candidate with the highest value $y^{4m+2}_k$. If $y^{4m+2}_k \geq 2$
        then $d_k$ is eliminated. This only increases the veto-score of $s_2$ by 1
        and does not affect the veto-scores of other running candidates.
        Suppose that there exists $k$ such that $y^{4m+2}_k = 0$. In this case $p$ has 1 veto-point extra compared to $d_k$.
        Moreover, the manipulator cannot save $p$ from elimination due to the tie-breaking rule.
        Hence, $y^{4m+2}_i \geq 2$ for  $i \in [1,n]$.
        This means that
        sets $S_j$ that correspond to  candidates $b_j$ that are eliminated during the first phase cover all values.
        Next we show that exactly $n/3$ of $b_j$'s are eliminated.
        %\item[The candidate $d_0$:] 
        \newline  \indent \textit{$\triangleright$ The candidate $d_0$:}
        This candidate has $2(m-n/3)+1$ veto-points less than the veto-score of $p$. Hence, during the first phase  $d_0$
        needs to get at least $2(m-n/3)$ extra veto-points. This means that $m-n/3$ of the candidates $\bar{b}_j$ have to be eliminated during the first phase.
        Hence exactly $n/3$  $b_j$'s can be eliminated during the first phase.         
        Finally, the manipulator gives one extra veto-point to $d_0$
        and, by the tie-breaking rule, $d_0$ is eliminated. Hence, $s_2$ is eliminated after $d_i$s, and  $p$ reaches the next round.
%The resulting veto-scores after $4m+2+(n+1)$ rounds are shown at Table~\ref{t:t1} (Round $4m+1+{n+1}$)
   %   \end{description}
       %Hence, if the candidates $b_j$ that are eliminated in the first phase correspond to sets that form a cover, the manipulator can ensure that   $p$ is not %eliminated during the first 3 phases.

       \textbf{Phase 4. Garbage collection. Rounds $4m+2+(n+1)+1$ to $c$.}
       This phase ensures that $p$ wins if it is not already eliminated.
        $p$ is either the last, first or second candidate in all remaining votes at this round.
       Hence, its veto-score does not change until the penultimate round.
The elimination of $d_0$, and then $s_2$, increases the veto-score of a candidate  $g_{d_0}$ by $2X- (n+1)-2(m-n/3)-5$ (lines 2 and 35),
which triggers elimination of other running candidates up to round $c-2$. 
    %As $d_0$ was eliminated, the  veto-score of $g_{d_0}$ is $X-2(m-n/3+1)-1$, hence the total  veto-score is $2X- (n+1)-2(m-n/3)-5$ which is %greater than the  veto-score of any other candidate. So $g_{d_0}$ is eliminated next. Then $g_{d_1}$ gets the same additional number of points %from line 35 and line 2. Therefore, $g_{d_1}$ now has the most veto-points and is eliminated next. This pattern of elimination continues until %2 remain. The candidate $p$ will not be eliminated because it cannot move to the last place in any more votes while more than 2 candidates remain. 
When only 2 candidates remain,  $p$ must win.

The reverse direction is trivial. Given a cover $I$, we construct the vote of a manipulator in the following way.
If $i$ is in cover, we put $a_i$ at position  $c - 2i$ and $\bar{a}_i$ at position $c - 2i -1$.
Otherwise, we invert their positions.
Then we put $d_0,s_2$. Finally, we make $p$ the most
preferred candidate, and put the remaining candidates
in an arbitrary order.
		\myqed

\section{Coombs' Rule}

Coombs' rule is a variant of $eliminate(veto)$
with the stopping criteria that a winner is declared
when one candidate has a majority
of first placed votes (instead of when one candidate
remains). Although this is a small change,
it can have a large impact on the result and
on strategic voting. For instance, there
are a family of elections where the number
of manipulators required to achieve
victory for a particular candidate
is unbounded for $eliminate(veto)$
but bounded for Coombs', and
vice versa.

\begin{mytheorem}
There exists an election with $n+3$ candidates
where a given candidate has already won with $eliminate(veto)$
but the number of manipulators with Coombs' rule is $\Omega(n)$.
\end{mytheorem}		
\myproof
We have $n$ votes:
$(a,d_1,\myldots,d_n,b,c),$
$(a,d_2,\myldots,d_{n-1},b,c),$
\myldots,
$(a,d_n,\myldots,d_1,b,c)$.
\nina{Note that positions $2-(n+1)$ in these votes
contain a cyclic permutation of candidates $d_1,\ldots,d_n$. 
Similarly, for the other two groups of $n$ votes.}
We also have $n$ votes:
$(b,a,d_1,\myldots,d_n,c)$,
$(b,a,d_2,\myldots,d_{n-1},c)$,
\myldots,
$(b,a,d_n,\myldots,d_1,c)$.
Finally we have $n$ votes:
$(c,b,a,d_1,\myldots,d_n)$,
$(b,a,d_2,\myldots,d_{n-1})$,
\myldots,
$(c,b,a,d_n,\myldots,d_1)$.
The preferred candidate is $a$.
As is common in the literature,
ties are broken in favor of manipulators.
For $eliminate(veto)$,
$c$ is eliminated in the first round and $b$ in the
second. $a$ is now in first place in all votes
so ultimately wins. 
For Coombs', $c$ is eliminated in the first round.
$b$ is then in the first place in $2n$
votes and wins by the majority rule. There are two options for
the manipulators. Either they add $n$ votes to the elections
to make sure that $b$ does not have a majority after the first
elimination round, or they prevent the elimination of the candidate
$c$ in the first round. With the exception of $c$, each candidate has
has only one veto point. Therefore, the manipulators need 
at least $2n-1$ votes to prevent the elimination of $c$.
\myqed

\begin{mytheorem}
		%\label{l:l1}
There exists an election with $n+2$ candidates
which a single manipulator can manipulate
with Coombs' rule but 
$eliminate(veto)$ requires $\Omega(n)$ manipulators.
\end{mytheorem}
\myproof
We have $n$ votes:
$(a,b,d_1,\myldots,d_n),$
$(a,b,d_2,\myldots,d_{n-1}),$
\myldots,
$(a,b,d_n,\myldots,d_1)$.
\nina{Note that positions $2-(n+1)$ in these votes
contain a cyclic permutation of candidates $d_1,\ldots,d_n$. 
Similarly, for the other group of $n$ votes.}
We also have $n$ votes:
$(b,d_1,\myldots,d_n,a)$,
$(b,d_2,\myldots,d_{n-1},a)$,
\myldots,
$(b,d_n,\myldots,d_1,a)$.
None of the candidates has a majority.
For Coombs', if one manipulator puts $a$ in first place then
$a$ wins.
For $eliminate(veto)$,
the manipulators must prevent the elimination of $a$ in the
first round. As candidates $d_1$ to $d_n$
have only 1 veto point we need at least $n-1$ manipulators
to prevent the elimination of $a$.
\myqed

Despite these differences between Coombs' rule
and $eliminate(veto)$, it is intractable
to compute a manipulation for Coombs' as it
is with $eliminate(veto)$.

\begin{mytheorem}
		\label{l:l2}
        Deciding if a single manipulator can make a candidate
win for the Coombs' rule is NP-complete.
		\end{mytheorem}		
		\myproof
Follows from the  proof of %reduction in
 Theorem~\ref{l:l1}
as  $g'_i$ can be eliminated after a cover is verified.
%We use the same reduction as in the proof of Theorem~\ref{l:l1}.
%A majority winner cannot emerge till the fourth phase.
%If a manipulator can make his preferred candidate $p$
%win the election then
%the first  garbage collector is removed during the fourth phase {after a
%cover is verified}. The elimination
%of a  garbage collector increases the plurality score of $p$ only.
%Hence, when enough garbage collectors are removed, $p$ becomes a  plurality winner.
\myqed

\section{Eliminate(scoring rule)}

We next consider scoring rules in
general. With weighted votes, Coleman and Teague (\citeyear{ctcats2007})
have proved that
manipulation by a coalition is NP-hard
to compute for the elimination version
of any scoring rule $X$ that is not isomorphic to veto.
% and the number of candidates is bounded
%\cite{ctcats2007}.
%We consider elimination versions of
%scoring rules with unweighted votes.
With unweighted votes, 
we prove a general result that relates the 
computational complexity 
of manipulating a scoring rule and 
the elimination version of its adjoint.

\begin{mytheorem}
Deciding whether $k$ manipulators can make a candidate
win for $eliminate(X)$ is NP-complete
if it is NP-complete also for $X^*$. 
\label{thm-adjoint}
		\end{mytheorem}		
		\myproof
First, we argue that for votes $V$, $k$ manipulators
can make a candidate win with $X^*$ iff 
for the reversed set of votes $V^*$, $k$ manipulator can make a
candidates come last with $X$. The
proof is similar to Lemma 10 in \cite{ctcats2007}. We 
simply reverse all the manipulating votes.
Suppose $V^*$ is an election over $m$ candidates
where $m \geq 3$,
and the $k$ manipulators want $c_m$ 
to come last. Let $U$ be 
$s_1(|V| + k + 1)$ copies of the following votes: 
\begin{center}
\begin{tabular}{c}
$c_1 \succ c_2 \succ \ldots \succ c_{m-1} \succ c_m$, \\
$c_2 \succ c_3 \succ \ldots \succ c_m \succ c_1$, \\
%$c_3 \succ c_4 \succ \ldots \succ c_1 \succ c_2$, &
$\vdots$ \\ % &
$c_m \succ c_{1} \succ \ldots \succ c_{m-2} \succ c_{m-1}$\\
\end{tabular}
\end{center}
Each candidate receives the same score in $U$
irrespective of $X$. We argue that 
there is a manipulation making 
$c_1$ win in $V^* \cup U$ for $eliminate(X)$
if there is a manipulation making 
$c_m$ last in $V^*$ for $X$. 
By the same argument as in the proof of Theorem 13 in 
\cite{ctcats2007}, if $c_i$ is the first candidate 
eliminated in $V^* \cup U$, then no matter how the 
manipulators vote, the elimination order is
$c_i,c_{i-1},c_{i-2},\ldots, c_{i+1}$ (where $c_{m+1} = c_{1}$) 
and $c_{i+1}$ wins. Hence $c_1$ wins iff $c_m$ is eliminated first. 
The manipulators can force $c_m$ to be eliminated 
first in $V^*\cup U$ if they can force $c_m$ to be last in 
$V^*$ as the relative scores 
of the candidates are initially the same in  $V^*$ and in $V^*\cup U$. 
Hence manipulation of
$X^*$ reduces to manipulation of $eliminate(X)$. 
\myqed

Borda is NP-hard to manipulate with 2 manipulators
\cite{borda2,dknwaaai11}. Since the adjoint
of Borda is Borda itself, 
it follows from Theorem \ref{thm-adjoint} that 
$eliminate(Borda)$, which is
Baldwin's rule, is NP-hard to manipulate by 2 manipulators. 
This result is strengthened
to NP-hard with just one manipulator in \cite{nwxaaai11}. 
Note that the reverse of Theorem \ref{thm-adjoint}
does not hold. STV, which is $eliminate(plurality)$, is NP-hard
to manipulate but $plurality$ is only polynomial to
manipulate. 

We next identify a large class of
scoring rules which are intractable
to manipulate. 
Given a fixed $k$, a {\em truncated scoring rule}
(\truncated) has a scoring vector $(s_1,\ldots,s_m)$ with
$s_i=0$ for all $i>k$.
For example, plurality and $k$-approval voting are both truncated
scoring rules.
As a second example, the Heisman Trophy, 
which is awarded annually to the best player in collegiate football,
uses the truncated scoring rule
$(3,2,1,0,\ldots,0)$.
%As a third example, the presidential election in Kiribat
%uses the truncated scoring rule
%$(4,3,2,1,0,\ldots,0)$.
As a third and final example, the Eurovision song contest
uses the truncated
scoring rule $(12, 10, 8, 7, 6, 5, 4, 3, 2, 1, 0,\ldots,0)$.
We now prove out next major results:
computing a manipulation of
$eliminate(\truncated)$ or
of $divide(\truncated)$ is intractable.
When candidates are eliminated, we suppose that the
scoring vector is 
truncated to the first $m$ positions
where $m$ is the number of candidates left after elimination.

\begin{mytheorem}
        Deciding if a single manipulator can make a candidate
win for $eliminate(\truncated)$ is NP-complete.
		\end{mytheorem}		
		\myproof
(Sketch, the full proof can again be found online in a technical report).
{The proof is also inspired by ideas from~\cite{stvhard}
and uses a reduction from 3-COVER. 
We block
the first $k-1$ positions in each vote with an
additional set of $q(k-1)$ dummy candidates,
where the value $q$ is computed taking into account the scoring vector.
By this construction, only those
scores at positions $k$ to $c$, which are $(s_k,0,\ldots,0)$,
determine the elimination order for the first $c-q(k-1)-1$ rounds,
where $c$ is the total number of candidates.
We thereby reduce our problem to one that resembles
a multiple of the reduction used for STV.
Only one non-dummy candidate reaches the $(c-q(k-1)-1)$th round.
If the preferred candidate $p$ reaches this
round then the remaining votes are such that
$p$ wins the election.
Similar to the reduction used
in the STV proof, this only happens if there is a 3-COVER.
As we have a large number of additional dummy candidates, we can
make sure that the individual score of each dummy candidate is
greater than the score of any non-dummy candidate until
the $(c-q(k-1)-1)$th round and is
smaller than the score of $p$ at the $(c-q(k-1)-1)$th round.}
\myqed

\begin{mytheorem}
        Deciding if a single manipulator can make a candidate
win for $divide(\truncated)$ is NP-complete.
		\end{mytheorem}		
		\myproof
(Sketch, the full proof can again be found online in a technical report).
{
The proof uses a reduction from the 3-COVER problem where $k=n/3$,
$n$ is the number of items.
The first two rounds encode solving the 3-COVER problem and the remaining rounds are used to collect garbage.
The main types of candidates are $n$ `items',  $m$ `sets' and one `preferred'.
The rest of the candidates are dummy candidates that are used to control
scores of non-dummy candidates and the average score.
In the first round, we select $k$ sets. 
Using a large number of dummy candidates we make sure 
that the score of 'sets' candidates equals
the average score at the first round. Hence, manipulator
can select $k$ of them to pass to the second round.}
In the second round, we check that this forms a cover. If
this is the case, all 'items' candidates 
in the covered set are eliminated. Otherwise, one of the `items'
candidates reaches the third round and wins the election.
%In the remaining round, all other candidates but the
%preferred candidate are eliminated.
\myqed

\section{Sequential Rules}

When selecting the site of the next Olympics, IOC
members can cast a new vote in each round.
This increases the opportunity for manipulation.
In fact, we can exhibit an election in which a manipulator
can only change the result
if the manipulator votes differently in each round.

\begin{myexample}
Consider the following $21$ votes:
\begin{tabular}{lll}
1: (a, h, p, \ldots), & 1: (c, a, h, p, \ldots), & 1: (d, a, h, p, \ldots) \\
3: (g, a, h, p, \ldots), & 2: (b, h, p, \ldots), & 2: (e, b, h, p, \ldots) \\
2: (f, b, h, p, \ldots), & 6: (h, p, \ldots), & 5: (p, h, \ldots)
\end{tabular}
The election uses $sequential(plurality)$,
and the manipulator wants $p$ to win.
The tie-breaking rule is $p \prec g \prec c \prec d \prec a \prec e \prec f \prec b \prec h$.

We first argue that 
a single manipulator cannot make $p$ win.
Note that $p$ cannot gain any points until $h$ is eliminated.
%as $p$ appears after $h$ in every vote where $p$
%is not already in first place. 
For $p$ to win, the manipulator
needs to give $p$ one point so that it has 6 points (and beats $h$ 
by the tie-breaking rule) and no other candidate receives more 
than $6$ points. In order for $h$ not to receive any
more than the initial 6 points, $a$ and $b$ must not be eliminated.
The manipulator must save $a$ from elimination in the first round by voting for $a$. The first two rounds therefore eliminate $c$ and $d$.
%, after which $a$ has $3$ points and is safe for now. After the first two rounds, 
Unfortunately, the manipulator cannot stop $b$ being eliminated next. 
$h$ now receives $2$ more points and $p$ cannot win the election.
Therefore, a single manipulator cannot make $p$ win.

On the other hand, if a single manipulator votes for $a$ in the first two rounds, $c$ and $d$ are eliminated and $a$ has $3$ points and is safe for now. At this point, $b$ is in danger with only $2$ points. If the manipulator now votes for $b$, $b$ is saved and both $e$ and $f$ are eliminated next. At this point, $a$ and $g$ are tied. If the manipulator now votes for $a$ again, $g$ is eliminated and the score of $a$ increases to $6$. At this point, all candidates except $p$ have $6$ points, and if the manipulator now votes for $p$, $h$ is eliminated by tie-breaking and $p$ wins the election. Hence, if the manipulator can change votes after each round, $p$ can win.
\end{myexample}

In general, manipulating a sequential elimination election
requires a strategy, which provides a manipulating
response however the other agents vote. It is not
hard to see that deciding if such a strategy
exists is PSPACE-complete.
In fact, strategic voting in a sequential elimination
election invites a game theoretic analysis.
We can view a sequential
elimination election as a finite repeated sequential game.
We could, for example,
use backward induction to find the subgame perfect
Nash equilibrium in which each agent makes the best strategic
move in each round.

\section{Other Related Work}

Bag, Sabourian and Winter (\citeyear{weakestlink})
proved that many
sequential elimination rules including $sequential(plurality)$
elect candidates in the top cycle (and
are hence Condorcet consistent)
supposing strategic voting. % \cite{weakestlink}.
%They illustrated this class %of sequential elimination rules
%with $sequential(plurality)$, which they call the
%``weakest link'' rule.
%
%Geller (\citeyear{geller}) has proposed a variant of STV
%which successively eliminates candidates based on their
%{\em original} Borda score. % \cite{geller}.
%Unlike Nanson's
%and Baldwin's rules, this method does not recalculate the Borda
%score based on the new reduced set of candidates.
%
%With a domain restriction,
%some of the resistance to manipulation
%identified for elimination
%style voting rules may disappear. For instance,
%for single peaked votes and a Condorcet consistent rule (and thus 
%Nanson's and Baldwin's rule),
%Brandt {\it et al.} 
%(\citeyear{bbhhaaai10}) showed that many types of
%control and manipulation problems are polynomial
%to compute.
%
Contizer and Sandholm 
(\citeyear{csijcai03}) studied the impact on the
tractability of manipulation
of adding an initial round of the Cup rule to a voting
rule.  
This initial round eliminates half the candidates and
makes manipulation NP-hard to compute
for several voting rule including plurality and Borda.
Elkind and Lipmaa 
(\citeyear{elisaac05}) extended this idea to
a general technique for combining
two voting rules.
The first voting rule is run for some number
of rounds to eliminate some of the candidates,
before the second voting rule is applied to
the candidates that remain. They proved that
many such combinations of voting rules are NP-hard
to manipulate. However, they did not consider the
veto or truncated scoring rules at the centre
of our study here. They also considered
the {\em closed protocol}, where a rule
is combined with itself. In many cases,
the closed protocol of $X$ is $eliminate(X)$.
They conjectured that such closed protocols will
often be NP-hard to manipulate.

%More recently, Xia and Walsh 
%(\citeyear{wxaamas12}) showed that
%using a lottery to eliminate some of the voters
%(instead of some of the candidates) is another
%route to make manipulation intractable to compute.
%%Adding such a pre-round to
%voting rules like Borda makes it NP-hard
%to compute a manipulation (as well as to
%compute the probability of a candidate winning).

\section{Conclusions}

We have provided more evidence that successively
eliminating candidates is often a route to
making manipulation intractable
to compute. In general, eliminating candidates does not
necessarily increase the computational complexity
of manipulation. Indeed, we exhibited an artificial
voting rule where the computational complexity actually
decreases. However, for many voting rules used in
practice, the computational complexity increases.
For example, it was known that it
is NP-hard to compute how a single voter
can manipulate the result of STV (the
elimination verison of plurality voting), and Nanson's
and Baldwin's rule (elimination versions
of Borda voting). In this paper,
we showed that it
is NP-hard to compute how a single voter
can manipulate the result of the elimination version
of veto voting, of the closely related Coombs' rule,
and of the elimination versions of a general class of
truncated scoring rules.
On the other hand, we showed that permitting voters
to re-vote between elimination rounds can
increase the opportunity for manipulation.

What general lessons can be learnt from these
studies? First, elimination style voting does indeed
appear to provide some computational resistance to
manipulation. 
%It would be worth considering
%other elimination style voting rules like
%$eliminate(approval)$ and $eliminate(maximin)$.
Second, these results have involved
worst case complexity notions like NP-hardness.
We need to treat these with care
as there is theoretical evidence (for instance,
\cite{xcec08,fknfocs09,xcec08b}),
as well as practical experience 
which suggests that elimination style
rules like STV
\cite{ctcats2007,wecai10}, as well as voting rules like
veto \cite{wijcai09} can be easy to manipulate on
average. 
Third, if voters can re-vote between elimination
rounds, new opportunities for manipulation arise.
It would be interesting therefore to consider
both game-theoretic and
computational aspects of such strategic voting.
For example, what are the possible equilibria
and how difficult are they to compute?
Fourth, manipulation is closely connected
to questions about possible winners given
uncertainty about the votes \cite{prvwijcai2007,waaai2007}
and to elicitation \cite{waamas08}. It would therefore be
interesting to consider reasoning about
possible winners and preference elicitation
for elimination style voting rules.

\section*{Acknowledgments}

The authors are supported by 
the Australian Government's Department of Broadband, Communications and
the Digital Economy, the Australian Research Council
and the Asian Office of Aerospace
Research and Development through grants AOARD-104123
and 124056.

\bibliographystyle{aaai}

%\bibliography{/Users/twalsh/Documents/biblio/a-z,/Users/twalsh/Documents/biblio/a-z2,/Users/twalsh/Documents/biblio/pub,/Users/twalsh/Documents/biblio/pub2}
\bibliography{/Users/twalsh/Documents/biblio/a-z2,/Users/twalsh/Documents/biblio/pub2}
%\bibliography{a-z2,pub2}
%\bibliography{/home/tw/biblio/a-z,/home/tw/biblio/a-z2,/home/tw/biblio/pub,/home/tw/biblio/pub2}

\end{document}